\documentclass[11pt, a4paper, logo, classicmath]{googledeepmind}

\usepackage[authoryear, sort&compress, round]{natbib}
\usepackage{mathtools}
\usepackage{dsfont}
\usepackage{longtable}
\usepackage{tikz}
\usetikzlibrary{arrows.meta}
\usepackage[capitalise, noabbrev]{cleveref}
\crefname{appsec}{Appendix}{Appendices}

\titlespacing*{\paragraph}{0pt}{1.2ex plus 0.4ex minus 0.2ex}{1em}
\theoremstyle{plain}
\newtheorem{theorem}{Theorem}
\newtheorem{lemma}{Lemma}
\newtheorem{proposition}{Proposition}
\newtheorem{corollary}{Corollary}
\theoremstyle{definition}
\newtheorem{definition}{Definition}
\theoremstyle{remark}
\newtheorem{remark}{Remark}

\newcommand{\R}{\mathbb{R}}
\newcommand{\Q}{\mathbb{Q}}
\newcommand{\Z}{\mathbb{Z}}
\newcommand{\N}{\mathbb{N}}
\newcommand{\A}{\mathbb{A}}
\newcommand{\Qpi}{\mathbb{Q}_{p_i}}
\newcommand{\Zpi}{\mathbb{Z}_{p_i}}
\newcommand{\ind}{\mathds{1}}
\newcommand{\cB}{\mathcal{B}}
\newcommand{\cC}{\mathcal{C}}
\newcommand{\cF}{\mathcal{F}}
\newcommand{\cP}{\mathcal{P}}
\newcommand{\cU}{\mathcal{U}}
\newcommand{\cX}{\mathcal{X}}
\newcommand{\cY}{\mathcal{Y}}
\newcommand{\bx}{\mathbf{x}}

\newcommand{\bz}{\mathbf{z}}
\newcommand{\bu}{\mathbf{u}}
\newcommand{\bc}{\mathbf{c}}
\newcommand{\bb}{\mathbf{b}}
\newcommand{\bp}{\mathbf{p}}
\newcommand{\bk}{\mathbf{k}}
\newcommand{\bs}{\mathbf{s}}
\newcommand{\bgamma}{\boldsymbol{\gamma}}
\newcommand{\hz}{\widehat{\mathbf{z}}}
\newcommand{\hp}{\widehat{\mathbf{p}}}

\newcommand{\Kp}{K_{\mathbf{p}}}
\newcommand{\KR}{K_{\mathbb{R}}}
\newcommand{\norm}[1]{\lVert #1 \rVert}
\newcommand{\abs}[1]{\lvert #1 \rvert}
\newcommand{\frp}[1]{\{#1\}_p}

\let\oldchi\chi
\DeclareRobustCommand{\chi}{{\mathpalette\raisechi\relax}}
\newcommand{\raisechi}[2]{\raisebox{\depth}{$#1\oldchi$}}

\title{Adelic Neural Network Architectures via Universal Approximation}

\author[1]{Maxime Guigon}
\author[1]{Lucas Dixon}
\author[2]{Julien Randon-Furling}
\author[1]{Michaël E. Sander}

\affil[1]{Google DeepMind}
\affil[2]{Centre Borelli, ENS Paris-Saclay}
\correspondingauthor{guimax@google.com}
\renewcommand{\today}{November 2026}

\begin{abstract}
Real and $\bp$-adic spaces offer complementary inductive biases for machine learning: $\R$ naturally represents smooth variations of magnitude, whereas the ultrametric tree structure of $\Q_p$ naturally captures hierarchies and modular arithmetic. 
Combining both in a single neural network is not straightforward because $\R$ is connected whereas $\Q_p$ is totally disconnected, so naive cross-field layers either break continuity or discard the ultrametric geometry.
In this paper, we derive a new universal approximator on the proxi-adelic space $\cX \coloneqq \R^{d_0}\times\prod_{i=1}^{M}\Q_{p_i}^{d_i}$, which combines real coordinates with $\bp$-adic coordinates for finitely many distinct primes. 
For continuous maps on $\KR\times\prod_{i}\Z_{p_i}^{d_i}$, with $\KR$ compact and connected, we show that the topology of the space largely constrains the form of an approximator. 
A real-valued output is approximated by finitely many real networks, each switched on by the indicator of a ball in the $\bp$-adic coordinates. 
A $\mathbf{p_i}$-adic-valued output cannot depend on the real input at all, and depends on the other primes only through ball indicators. 
The result holds for any choice of universal real and $\bp$-adic approximators; instantiated with perceptrons and $\bp$-adic character networks, it yields an explicit universal layer. 
We then show that universality is preserved under two relaxations: replacing partitions by covers, which gives a successive refinement along the $\bp$-adic tree, and replacing ball indicators by their Fourier expansion, which turns the gating into a modulation by cosines of $\bp$-adic fractional parts.\end{abstract}

\begin{document}

\maketitle

\section{Introduction}
\label{sec:intro}

Real representations encode information as magnitudes along continuous directions, and similar inputs are those with nearby coordinates. 
For a prime $p$, the $p$-adic numbers $\Q_p$ measure closeness by divisibility: two numbers are close when their difference is divisible by a high power of $p$, i.e.\ when their base-$p$ expansions share a long common prefix. 
The resulting metric is ultrametric, its balls are either nested or disjoint, and $\Z_p$ is naturally organised as an infinite $p$-ary tree.
A $p$-adic representation thus encodes information as positions in a tree, and similar inputs are those that share a deep common ancestor.
These two notions of similarity are therefore very different.
For instance, linear models over $\Q_p$ can learn modular arithmetic, an XOR-like task that linear models over $\R$ cannot express \citep{salazar2026continuous}. 
This geometry is reminiscent of hierarchies and of modular arithmetic and could help machine learning systems represent such structure. 
We refer to \citet{martins2025learning} and \citet{salazar2026continuous} for general introductions to $p$-adic machine learning.
Conversely, smooth variations of magnitudes seem more natural over $\R$ than over $\Q_p$. 
Since these two geometries capture complementary aspects of data, the most expressive models may come from combining both representations within the same architecture.
\paragraph{An adelic architecture, and why it is not obvious.}
By Ostrowski's theorem, the only non-trivial completions of $\Q$ are $\R$ and the fields $\Q_p$. The adeles (\ref{sec:background}) gather all of them in a single locally compact ring. This gives the intuition that an adelic space could be a natural home for more general machine learning, in which every completion, and hence every kind of representation above, is available at once.
If such a combination is desirable, a first question is architectural: what should an adelic layer look like? The answer is not obvious, because the two topologies are hard to reconcile. $\R$ is connected while $\Q_p$ is totally disconnected, so for example every continuous map from a connected real set to $\Q_p$ is constant. Naive constructions, such as concatenating coordinates or embedding $p$-adic inputs into $\R$, may thus discard the structure we want to keep, or allow interactions that are not continuous. Rather than guessing an architecture, we look for one suggested by a universal approximation theorem.

\begin{figure}[htbp]
\centering
\begin{tikzpicture}[
    >=Latex,
    realC/.style={draw=blue!70!black, fill=blue!8},
    padC/.style={draw=red!70!black, fill=red!8},
    pOneC/.style={draw=red!70!black, fill=red!8},
    pTwoC/.style={draw=green!50!black, fill=green!8},
    io/.style={circle, thick, minimum size=1.05cm, inner sep=1pt, align=center, font=\small},
    approx/.style={rectangle, rounded corners=4pt, thick, minimum height=0.9cm, minimum width=2.6cm, align=center, font=\footnotesize},
    sel/.style={rectangle, thick, draw=orange!80!black, fill=orange!12, minimum height=0.8cm, align=center, font=\footnotesize},
    prod/.style={circle, thick, draw=gray!80!black, fill=gray!15, minimum size=0.75cm, inner sep=0pt},
    flow/.style={->, thick, draw=black!80},
    idx/.style={->, thick, dashed},
    lab/.style={font=\scriptsize, midway, above},
    dimlab/.style={font=\scriptsize},
]
    \node[io, realC, label={[dimlab]below:$\R$}]   (x) at (-3.5, 2.6)  {$x$};
    \node[io, padC,  label={[dimlab]below:$\Q_p$}] (z) at (-3.5, -0.6) {$z$};

    \node[approx, realC] (mlpR)  at (-1, 2.6) {$\mathbf{w}_U^{\top}\sigma(\mathbf{a}_Ux+\bb_U)$};
    \node[sel]           (selR)  at (-1.1, 1.0) {$\ind_{\Z_p}\big(p^{-s_U}(z-c_U)\big)$};
    \node[prod]          (prodR) at (1.5, 2.6) {$\otimes$};
    \node[io, realC, label={[dimlab]below:$\R$}] (F) at (3.45, 2.6) {$\hat F$};

    \node[approx, padC] (mlpP) at (0.0, -0.6) {$\boldsymbol{\lambda}^{\top}\chi(\mathbf{a}z+\bb)$};
    \node[io, padC, label={[dimlab]below:$\Q_p$}] (G) at (3.45, -0.6) {$\hat G$};

    \draw[flow] (x) -- (mlpR);
    \draw[flow] (mlpR) -- (prodR);
    \draw[flow] (prodR) -- node[lab] {$\sum_{U\in\cP}$} (F);
    \draw[idx, red!70!black] (z) -- (selR);
    \draw[flow] (selR) -- (prodR);

    \draw[flow] (z) -- (mlpP);
    \draw[flow] (mlpP) -- (G);

    \node[io, pOneC, label={[dimlab]below:$\Q_{p_1}$}] (z1) at (5.05, 2.6)  {$z_1$}; %
    \node[io, pTwoC, label={[dimlab]below:$\Q_{p_2}$}] (z2) at (5.05, -0.6) {$z_2$};

    \node[approx, pOneC] (mlp1)  at (7.90, 2.6) {$\boldsymbol{\lambda}_U^{\top}\chi_1(\mathbf{a}_Uz_1+\bb_U)$};
    \node[sel]           (sel1)  at (7.95, 1.5) {$\ind_{\Z_{p_2}}\!\big(\tfrac{z_2-c_U}{p_2^{s_U}}\big)$};
    \node[prod]          (prod1) at (10.3, 2.6) {$\otimes$};
    \node[io, pOneC, label={[dimlab]below:$\Q_{p_1}$}] (G1) at (12.0, 2.6) {$\hat G_1$};

    \node[approx, pTwoC] (mlp2)  at (7.90, -0.6) {$\boldsymbol{\lambda}_U^{\top}\chi_2(\mathbf{a}_Uz_2+\bb_U)$};
    \node[sel]           (sel2)  at (7.95, 0.5) {$\ind_{\Z_{p_1}}\!\big(\tfrac{z_1-c_U}{p_1^{s_U}}\big)$};
    \node[prod]          (prod2) at (10.3, -0.6) {$\otimes$};
    \node[io, pTwoC, label={[dimlab]below:$\Q_{p_2}$}] (G2) at (12.0, -0.6) {$\hat G_2$};

    \draw[flow] (z1) -- (mlp1);
    \draw[flow] (mlp1) -- (prod1);
    \draw[flow] (prod1) -- node[lab] {$\sum\limits_{U\in\cP_1}$} (G1);
    \draw[idx, green!50!black] (z2) -- (sel1.west);
    \draw[flow] (sel1.east) -- (prod1);

    \draw[flow] (z2) -- (mlp2);
    \draw[flow] (mlp2) -- (prod2);
    \draw[flow] (prod2) -- node[lab] {$\sum\limits_{U\in\cP_2}$} (G2);
    \draw[idx, red!70!black] (z1) -- (sel2.west);
    \draw[flow] (sel2.east) -- (prod2);

    \draw[gray!70, rounded corners] (-4.0, -1.85) rectangle (12.65, -3.55);
    \node[circle, thick, realC, minimum size=0.35cm] at (-3.3, -2.3) {};
    \node[dimlab, anchor=west] at (-2.9, -2.3) {real coordinate};
    \node[circle, thick, pOneC, minimum size=0.35cm] at (0.9, -2.3) {};
    \node[dimlab, anchor=west] at (1.2, -2.3) {$p$- or $p_1$-adic coordinate};
    \node[circle, thick, pTwoC, minimum size=0.35cm] at (5.2, -2.3) {};
    \node[dimlab, anchor=west] at (5.5, -2.3) {$p_2$-adic coordinate};
    \node[prod, minimum size=0.45cm, font=\scriptsize] at (9.1, -2.3) {$\otimes$};
    \node[dimlab, anchor=west] at (9.4, -2.3) {product};
    \node[rectangle, rounded corners=2pt, thick, draw=black!60, fill=black!5, minimum width=0.5cm, minimum height=0.3cm] at (-3.3, -3.1) {};
    \node[dimlab, anchor=west] at (-2.9, -3.1) {base approximator};
    \node[rectangle, thick, draw=orange!80!black, fill=orange!12, minimum width=0.5cm, minimum height=0.3cm] at (0.9, -3.1) {};
    \node[dimlab, anchor=west] at (1.2, -3.1) {ball indicator};
    \draw[flow] (4.95, -3.1) -- (5.45, -3.1);
    \node[dimlab, anchor=west] at (5.5, -3.1) {feature flow};
    \draw[idx, black!80] (8.85, -3.1) -- (9.35, -3.1);
    \node[dimlab, anchor=west] at (9.4, -3.1) {input of an indicator};

\end{tikzpicture}
\caption{\textbf{How topology shapes an adelic layer (\cref{cor:explicit}).} Within each field, coordinates are processed by native universal approximators (real perceptrons $\sigma$ in blue; $p$-adic character networks $\chi$ in red and green), while interactions across fields occur exclusively through $p$-adic ball indicators (orange), summed over a finite ball partition $\cP$. \emph{Left ($\R\times\Q_p$):} The $p$-adic coordinate $z$ gates a mixture of real networks on $x$ to form $\hat F$, whereas $x$ cannot reach the $p$-adic output $\hat G$ because every continuous map from $\R$ to $\Q_p$ is constant (\cref{cor:no-real}). \emph{Right ($\Q_{p_1}\times\Q_{p_2}$):} Between distinct primes, the coupling is symmetric: each coordinate $z_k$ gates a mixture of $p$-adic networks on the other coordinate $z_i$.}
\label{fig:arch-RxQp}
\end{figure}
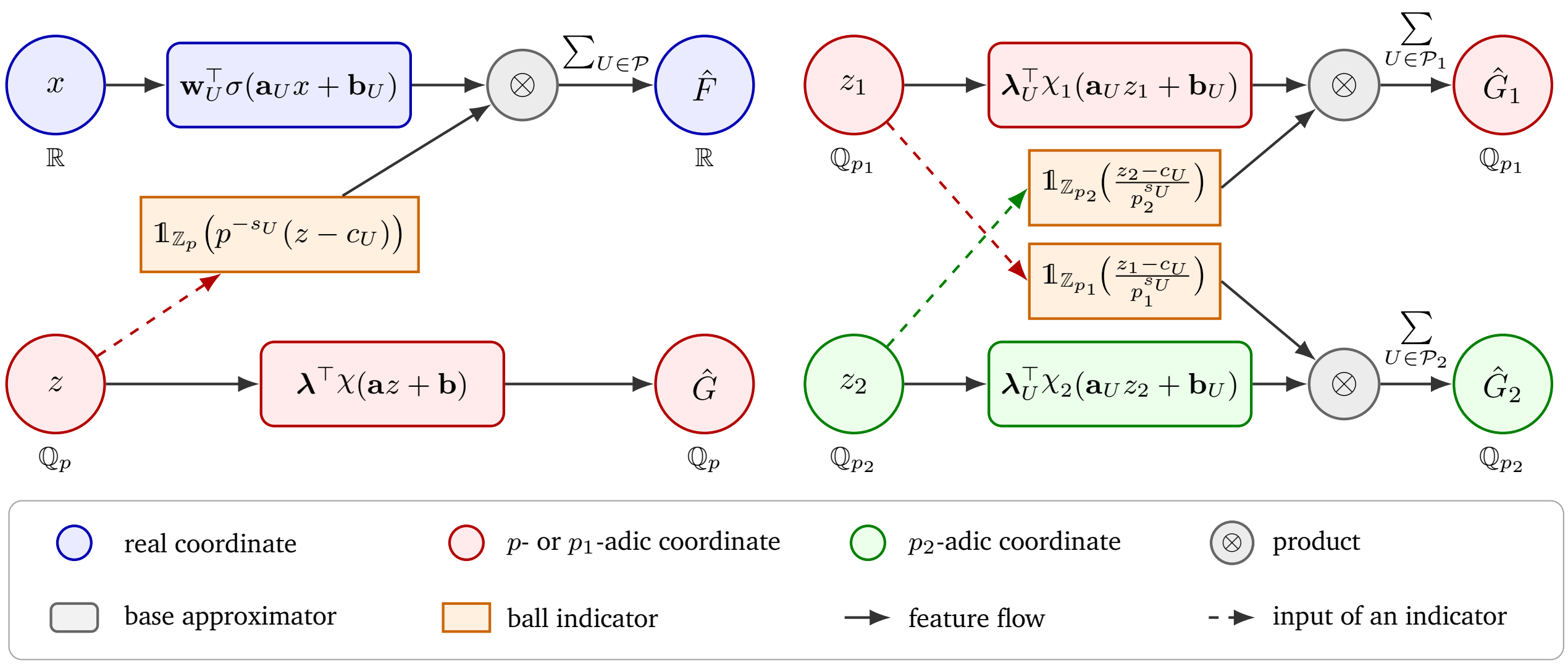

\paragraph{Contributions.}
We study continuous maps on compact domains $K = \KR \times \prod_{i=1}^M \Zpi^{d_i}$ of the proxi-adelic space $\cX = \R^{d_0} \times \prod_{i=1}^M \Q_{p_i}^{d_i}$ (with $\KR \subset \R^{d_0}$ compact and connected) and make the following contributions:
\begin{itemize}
  \item \textbf{A proxi-adelic universal approximation theorem (\cref{thm:main}).} Given \emph{any} universal approximator families $\cF_\R$ on $\KR$ and $\cF_{p_i}$ on $\Zpi^{d_i}$, we show that the product topology constrains how coordinates across different completions can interact:
  \begin{enumerate}[label=(\roman*)]
    \item \emph{Real-valued outputs} are uniformly approximated by finite sums of real networks in $\cF_\R$, each gated by the indicator of a ball from a finite partition of the $\bp$-adic domain (\cref{prop:real}).
    \item \emph{$p_i$-adic-valued outputs} cannot depend on the real input at all by connectedness (\cref{cor:no-real}), and are uniformly approximated by finite sums of $p_i$-adic networks in $\cF_{p_i}$ gated by ball indicators in the remaining primes $p_k \neq p_i$ (\cref{prop:padic}).
  \end{enumerate}
  \item \textbf{An explicit, computable adelic layer (\cref{sec:architecture,cor:explicit,fig:arch-RxQp}).} Instantiating the theorem with real perceptrons \citep{cybenko1989approximation}, $p$-adic character networks \citep{mihara}, and closed-form ball indicators yields a concrete universal architecture. Structurally, it acts as a \emph{$p$-adic-routed mixture of experts}, where the first $s_i$ digits of the $p_i$-adic inputs (congruence classes modulo $p_i^{s_i}$) select which network is evaluated.
  \item \textbf{Two flexible universal relaxations (\cref{sec:covers,sec:fourier}).} Because disjoint ball partitions induce hard routing across an exponentially growing number of unshared experts, we introduce two relaxations that preserve universality: (i) replacing partitions by overlapping \emph{ball covers} (\cref{sec:covers}), yielding a successive refinement along the $p$-adic tree; and (ii) replacing ball indicators by their \emph{Fourier expansion} (\cref{sec:fourier,prop:cos}), which turns hard gating into a soft modulation by cosines of $p$-adic fractional parts.
\end{itemize}
The appendix contains a self-contained treatment of the case $\R\times\Q_p$ with simple notation (\cref{app:demo}), which we recommend reading first as it captures the essence of the proof.

\section{Background}
\label{sec:background}

\paragraph{$\mathbf{p}$-adic numbers.}
For a prime $p$, $\abs{\cdot}_p$ denotes the $p$-adic absolute value and $\Q_p$ the completion of $\Q$ for it. It satisfies the ultrametric inequality $\abs{x+y}_p\le\max(\abs{x}_p,\abs{y}_p)$, and it takes value in $\{0\} \cup p^{\Z}$.
Every $x\in\Q_p\setminus \left\{ 0 \right\} $ has a unique expansion $x=\sum_{n\ge n_0}a_np^n$ with digits $a_n\in\{0,\dots,p-1\}$ and $a_{n_{0}}\neq 0$, and its fractional part is $\frp{x}:=\sum_{n<0}a_np^n\in\Z[1/p]\cap[0,1)$. 
The ring of $p$-adic integers $\Z_p:=\{x:\abs{x}_p\le1\}$ is compact and open. 
On $\Q_p^d$ we define $\norm{\bz}_p=\max_k\abs{z_k}_p$, and write $\cB(\bc,r):=\{\bz:\norm{\bz-\bc}_p\le r\}$ for a ball. Every ball has a radius of the form $p^{-k}$ with $k\in\Z$. 
Finally, we recall three standard facts: 
(i) every point of a ball is a centre of that ball, and two balls are either disjoint or nested; 
(ii) balls are both open and closed; 
(iii) $\Q_p^d$ is totally disconnected, i.e.\ its only connected subsets are points. 
For a detailed introduction, see \citet{gouvea2020padic,pomerantz2020padic}.

\paragraph{Adeles.}
The adele ring $\A_\Q=\R\times\prod'_p\Q_p$ is the restricted product of all completions of $\Q$ with respect to the subrings $\Z_p$. For a finite set $S$ of primes, $\R\times\prod_{p\in S}\Q_p\times\prod_{p\notin S}\Z_p$ is an open subring. The space $\cX$ studied here (when setting $d_0=d_i=1$) is its first $|S|+1$ factors. This is why we call it proxi-adelic.

\paragraph{Universal approximation over $\R$.}
One-hidden-layer perceptrons $\bx\mapsto\sum_jw_j\sigma(\mathbf{a}_j\cdot\bx+b_j)$ are dense in $\cC(K,\R)$ for every compact $K\subset\R^d$, for continuous sigmoidal $\sigma$ \citep{cybenko1989approximation,hornik1989multilayer} and more generally for any continuous non-polynomial $\sigma$ \citep{leshno1993multilayer}. Applied coordinatewise, this gives density in $\cC(K,\R^n)$.

\paragraph{Universal approximation over $\Q_p$.}
$p$-adic neural networks go back to \citet{albeverio1999padic}. \citet{mihara} uses a single injective $p$-adic character $\chi:\Z_p\to\Q_p^\times$.

\begin{theorem}[\citealp{mihara}]\label{thm:mihara}
Let $f:\Z_p^d\to\Q_p^m$ be continuous and $\varepsilon>0$. There exist $h\in\N$, $A\in\Z_p^{h \times d}$, $\bb\in\Z_p^h$ and $C\in \Q^{m \times h}_p$ such that if $\chi$ is applied coordinatewise, $\sup_{\bz\in\Z_p^d}\norm{f(\bz)-C\chi(A\bz+\bb)}_p\le\varepsilon$.
\end{theorem}

\citet{kissandpal} determine the minimal width of universal $p$-adic ReLU networks for $L^q$ and $C^1$ norms. 
Those two results gave us computable architectures that form universal approximator families.

\paragraph{Optimisation over $\Q_p$.}
Since the value group of $\Q_p$, $\text{Im}(\abs{\cdot}_p) = \left\{ 0 \right\}\cup p^{\mathbb{Z}}$, is discrete, standard losses of $p$-adic parameters are flat away from their minima, and native optimisers have mostly been discrete searches. \citet{salazar2026continuous} propose continuous gradient descent for $p$-adic parameters by working on the Berkovich affine line, a path-connected metric tree that contains $\Q_p$ and whose points include the closed balls of $\Q_p$. 

\section{Notations}
\label{sec:notations}

Throughout, $\N=\{0,1,2,\dots\}$ includes $0$, both for dimensions and for exponents; a zero dimension gives the trivial space $\R^0=\Q_p^0=\Z_p^0=\{0\}$. Fix $d_0\in\N$, distinct primes $p_1,\dots,p_M$ and $d_1,\dots,d_M\in\N$. An index of notation is given in \cref{app:index}.

\paragraph{Spaces.}
Let $V=\prod_{i=1}^M\Q_{p_i}^{d_i}$ and $\cX=\R^{d_0}\times V$. Throughout, a point of $\cX$ is written $(\bx,\bz)$, with $\bx\in\R^{d_0}$ its real part and $\bz\in V$ its $p$-adic part. Elements of a product are written blockwise, $\bz=(\bz_1,\dots,\bz_M)$ with $\bz_i\in\Q_{p_i}^{d_i}$. We write $\hz_i=(\bz_1,\dots,\bz_{i-1},\bz_{i+1},\dots,\bz_M)\in\widehat V_i\coloneqq\prod_{k\ne i}\Q_{p_k}^{d_k}$ for $\bz$ without its $i$-th block, and $(\bz_i,\hz_i)$ for $\bz$ itself. We equip $V$, $\widehat V_i$ and $\cX$ with the max metrics
\[
  d_V(\bz,\bz')=\max_{1\le i\le M}\norm{\bz_i-\bz'_i}_{p_i},
  \qquad
  d\big((\bx,\bz),(\bx',\bz')\big)=\max\big(\norm{\bx-\bx'},\,d_V(\bz,\bz')\big),
\]
where $\norm{\cdot}$ is any norm on $\R^{d_0}$. For $\bs\in\Z^M$, we write $\bp^{\bs}\odot\bz=(p_1^{s_1}\bz_1,\dots,p_M^{s_M}\bz_M)$.

\paragraph{Balls.}
In a product of $p$-adic spaces with different primes, the natural analogue of a ball is an \emph{anisotropic} ball, with one radius per prime, as follows:

\begin{definition}[Balls of $V$]\label{def:balls}
For $\bc=(\bc_1,\dots,\bc_M)\in V$ and $\bs\in\Z^M$, the ball of centre $\bc$ and multi-radius $\bp^{-\bs}$ is
\[
  \cB_V(\bc,\bp^{-\bs})=\prod_{i=1}^M\cB(\bc_i,p_i^{-s_i}).
\]
\end{definition}
The balls of $V$ form a fundamental system of clopen neighbourhoods, and any point of a ball of $V$ is a centre of it. Balls of $\widehat V_i$ are defined in the same way with the primes $\hp_i=(p_k)_{k\ne i}$.

\paragraph{Compact domain.}
Let $\KR\subset\R^{d_0}$ be non-empty, compact and connected, and set
\[
  \Kp=\prod_{i=1}^M\Zpi^{d_i},\qquad
  K_{\hp_i}=\prod_{k\ne i}\Z_{p_k}^{d_k},\qquad
  K=\KR\times\Kp\subset\cX.
\]
When $M=1$, $K_{\hp_1}$ is an empty product, i.e.\ a single point. Note that we can work on the unit balls $\Zpi^{d_i}$ rather than arbitrary compact subsets of $\Q_{p_i}^{d_i}$ without loss of generality, because any compact subset of $V$ is bounded, so a linear rescaling $\bz_i\mapsto p_i^{s_i}\bz_i$ ($s_i\in\N$) maps it into $\Kp$. This choice also matches the canonical subrings defining $\A_\Q$ and the domain of standard $p$-adic approximators (\cref{thm:mihara}).

\paragraph{Outputs and uniform distance.}
Fix $n,n_1,\dots,n_M\in\N$ and $\cY=\R^n\times\prod_{i=1}^M\Q_{p_i}^{n_i}$ with the max metric $d_\cY$ (see \cref{app:different_primes} for different input and output primes). For topological spaces $A,B$ with $B$ metric, $\cC(A,B)$ is the set of continuous maps $A\to B$, with the uniform distance $d_\infty(F,G)=\sup_{a\in A}d_B(F(a),G(a))$. Since a map into a product is continuous if and only if its components are,
\[
  \cC(K,\cY)=\cC(K,\R^n)\times\prod_{i=1}^M\cC(K,\Q_{p_i}^{n_i}),
  \quad\text{and}\quad
  d_\infty(F,G)=\max\Big(d_\infty(F_0,G_0),\ \max_i d_\infty(F_i,G_i)\Big).
  \tag{1}\label{eq:product}
\]
Taking $n=d_0$ and $n_i=d_i$ gives $\cC(K,\cX)$. We denote by $\cF_\R$ any dense subset of $\cC(\KR,\R^n)$, and for each $i$, $\cF_{p_i}$ any dense subset of $\cC(\Zpi^{d_i},\Q_{p_i}^{n_i})$, each for the uniform distance.

\section{Universal approximation theorem}
\label{sec:results}

In this section we present and prove our main result which is stated first as \cref{thm:main}.

\begin{mdframed}[linewidth=0.8pt, linecolor=blue!50!black, backgroundcolor=blue!4, roundcorner=4pt, innertopmargin=4pt, innerbottommargin=8pt, innerleftmargin=8pt, innerrightmargin=8pt, nobreak=true]
  \begin{theorem}[Proxi-adelic universal approximation]\label{thm:main}
    Let
    \[
      S_\R=\left\{
        \begin{aligned}
          f\colon K&\longrightarrow\R^n\\
          (\bx,\bz)&\longmapsto\sum_{U\in\cP}\ind_U(\bz)\,f_U(\bx)
        \end{aligned}
        \ ,\ 
        \begin{aligned}
          &\cP\text{ a finite partition of $\Kp$ into balls: } \Kp=\bigsqcup_{U\in\cP}U\\
          &f_U\in\cF_\R
        \end{aligned}
        \right\},
        \]
        and, for $i\in\{1,\dots,M\}$,
        \[
          S_{p_i}=\left\{
            \begin{aligned}
              f\colon K&\longrightarrow\Q_{p_i}^{n_i}\\
              (\bx,\bz)&\longmapsto\sum_{U\in\cP_i}\ind_U(\hz_i)\,g_U(\bz_i)
            \end{aligned}
            \ ,\ 
            \begin{aligned}
              &\cP_i\text{ a finite partition of $K_{\hp_i}$ into balls: } K_{\hp_i}=\bigsqcup_{U\in\cP_i}U\\
              &g_U\in\cF_{p_i}
            \end{aligned}
            \right\}.
            \]
            
            $S_\R$ is dense in $\cC(K,\R^n)$ and, for every $i\in\{1,\dots,M\}$, $S_{p_i}$ is dense in $\cC(K,\Q_{p_i}^{n_i})$, for the uniform distance. Consequently,
            \[
              S_\R\times\prod_{i=1}^M S_{p_i}\ \text{is dense in}\ \cC(K,\cY),
            \]
            and in particular in $\cC(K,\cX)$ when $n=d_0$ and $n_i=d_i$.
            \end{theorem}
          \end{mdframed}

\noindent The proof of \cref{thm:main} follows from the following intermediate results, for which full proofs are in the appendix. The proof is a blockwise generalization of the $\R\times\Q_p$ case, treated with simpler notation in \cref{app:demo}, which contains all the essence of the proof and which we recommend to read first.

\begin{proposition}[Partition property]\label{prop:partition}
Let $k\in\N$ and $C_k=\{0,1,\dots,p^k-1\}^d$. The spaces $\Z_p$, $\Z_p^d$ and $\Kp$ can all be partitioned into finitely many balls of arbitrarily small size (proof in \cref{app:balls}):
\begin{enumerate}[label=(\roman*)]
  \item $\Z_p^d=\bigsqcup_{\bc\in C_k}\cB(\bc,p^{-k})$; in particular $\Z_p=\bigsqcup_{c=0}^{p^k-1}\cB(c,p^{-k})$.
  \item For $\bk\in\N^M$ and $C_\bk=\prod_{i=1}^M\{0,\dots,p_i^{k_i}-1\}^{d_i}\subset\Kp$, we have $\Kp=\bigsqcup_{\bc\in C_\bk}\cB_V(\bc,\bp^{-\bk})$.
  \item Two points in the same ball of (ii) are at distance at most $\max_ip_i^{-k_i}$ for the metric $d_V$.
\end{enumerate}
\end{proposition}

\begin{theorem}[Heine's theorem, {\citealp[Theorem~4.19]{rudin1976principles}}]
A continuous map from a compact metric space to a metric space is uniformly continuous. 
\end{theorem}

\begin{lemma}\label{lem:constant}
Let $X$ be a non-empty connected topological space and $g:X\to\Q_p^m$ be continuous. Then $g$ is constant (proof in \cref{app:compact}).
\end{lemma}

\begin{proposition}[The real-valued part]\label{prop:real}
$S_\R$ is dense in $\cC(K,\R^n)$.
\end{proposition}

\begin{proof}
Let $F\in\cC(K,\R^n)$ and $\varepsilon>0$. Since $K$ is a compact metric space (\cref{lem:compact}), Heine's theorem makes $F$ uniformly continuous: there is $\eta>0$ such that $d((\bx,\bz),(\bx',\bz'))\le\eta$ implies $\norm{F(\bx,\bz)-F(\bx',\bz')}\le\varepsilon/2$ for all $(\bx,\bz),(\bx',\bz')\in K$.

Choose $\bk\in\N^M$ with $p_i^{-k_i}\le\eta$ for all $i$, and let $\cP=\{\cB_V(\bc,\bp^{-\bk}):\bc\in C_\bk\}$ be the partition of $\Kp$ given by \cref{prop:partition}. For $U\in\cP$, denote its centre by $\bc_U\in\Kp$, and define $F_U:\KR\to\R^n$ by $F_U(\bx)=F(\bx,\bc_U)$. It is continuous. For $\bx\in\KR$ and $\bz\in U$,
\[
  d\big((\bx,\bz),(\bx,\bc_U)\big)=d_V(\bz,\bc_U)\le\max_i p_i^{-k_i}\le\eta,
  \quad\text{so}\quad
  \norm{F(\bx,\bz)-F_U(\bx)}\le\varepsilon/2.
\]
By density of $\cF_\R$, pick $f_U\in\cF_\R$ with $\sup_{\KR}\norm{F_U-f_U}\le\varepsilon/2$, and set $G(\bx,\bz)=\sum_{U\in\cP}\ind_U(\bz)f_U(\bx)$, so that $G\in S_\R$. For $(\bx,\bz)\in K$, exactly one $U\in\cP$ contains $\bz$, and $G(\bx,\bz)=f_U(\bx)$. Hence
\[
  \norm{F(\bx,\bz)-G(\bx,\bz)}\le\norm{F(\bx,\bz)-F_U(\bx)}+\norm{F_U(\bx)-f_U(\bx)}\le\varepsilon.
\]
Thus $d_\infty(F,G)\le\varepsilon$.
\end{proof}

\begin{corollary}[No real variable]\label{cor:no-real}
Every $F\in\cC(K,\Q_{p_i}^{n_i})$ is independent of $\bx$: for any $\bx_0\in\KR$, $F(\bx,\bz)=F(\bx_0,\bz)$ for all $(\bx,\bz)\in K$.
\end{corollary}

\begin{proof}
For all $\bz\in\Kp$, apply \cref{lem:constant} to $\bx\mapsto F(\bx,\bz)$ on the connected set $\KR$.
\end{proof}

\begin{proposition}[The $\mathbf{p}$-adic-valued parts]\label{prop:padic}
For every $i\in\{1,\dots,M\}$, $S_{p_i}$ is dense in $\cC(K,\Q_{p_i}^{n_i})$.
\end{proposition}

\begin{proof}
Let $F\in\cC(K,\Q_{p_i}^{n_i})$ and $\varepsilon>0$. Fix $\bx_0\in\KR$ and let $\tilde F=F(\bx_0,\cdot)\in\cC(\Kp,\Q_{p_i}^{n_i})$, so that $F(\bx,\bz)=\tilde F(\bz)$ on $K$ by \cref{cor:no-real}. The elements of $S_{p_i}$ do not depend on $\bx$ either, so it suffices to find $\tilde G(\bz)=\sum_{U\in\cP_i}\ind_U(\hz_i)g_U(\bz_i)$ with $\norm{\tilde F(\bz)-\tilde G(\bz)}_{p_i}\le\varepsilon$ for all $\bz\in\Kp$.

$\Kp$ is compact, so $\tilde F$ is uniformly continuous: there is $\eta>0$ such that $d_V(\bz,\bz')\le\eta$ implies $\norm{\tilde F(\bz)-\tilde F(\bz')}_{p_i}\le\varepsilon$. Choose $k_l\in\N$ with $p_l^{-k_l}\le\eta$ for $l\neq i$. \Cref{prop:partition}, applied to the $M-1$ primes $\hp_i$, gives a partition $\cP_i$ of $K_{\hp_i}$ into balls $U$ of centre $\bc_U\in K_{\hp_i}$ with $d_{\widehat V_i}(\hz_i,\bc_U)\le\eta$ whenever $\hz_i\in U$. When $M=1$, take $\cP_1=\{K_{\hp_1}\}$. For $U\in\cP_i$, define $\tilde F_U:\Zpi^{d_i}\to\Q_{p_i}^{n_i}$ by $\tilde F_U(\bz_i)=\tilde F(\bz_i,\bc_U)$, which is continuous. If $\hz_i\in U$, then
\[
  d_V\big((\bz_i,\hz_i),(\bz_i,\bc_U)\big)=d_{\widehat V_i}(\hz_i,\bc_U)\le\eta,
  \quad\text{so}\quad
  \norm{\tilde F(\bz)-\tilde F_U(\bz_i)}_{p_i}\le\varepsilon.
\]
By density of $\cF_{p_i}$, pick $g_U\in\cF_{p_i}$ with $\sup\norm{\tilde F_U-g_U}_{p_i}\le\varepsilon$, and let $\tilde G(\bz)=\sum_{U\in\cP_i}\ind_U(\hz_i)g_U(\bz_i)$. For $\bz\in\Kp$, exactly one $U$ contains $\hz_i$, and by the ultrametric inequality
\[
  \norm{\tilde F(\bz)-\tilde G(\bz)}_{p_i}\le\max\Big(\norm{\tilde F(\bz)-\tilde F_U(\bz_i)}_{p_i},\ \norm{\tilde F_U(\bz_i)-g_U(\bz_i)}_{p_i}\Big)\le\varepsilon.
\]
Hence $G(\bx,\bz)=\tilde G(\bz)$ belongs to $S_{p_i}$ and $\norm{F(\bx,\bz)-G(\bx,\bz)}_{p_i}=\norm{\tilde F(\bz)-\tilde G(\bz)}_{p_i}\le\varepsilon$ on $K$, i.e.\ $d_\infty(F,G)\le\varepsilon$.
\end{proof}

We can now prove \cref{thm:main}.

\begin{proof}[Proof of \cref{thm:main}]
The two density statements are \cref{prop:real,prop:padic}. For the product, let $\varepsilon>0$ and $F=(F_0,F_1,\dots,F_M)$ in $\cC(K,\cY)$, and pick $G_0\in S_\R$ and $G_i\in S_{p_i}$ that are $\varepsilon$-close to $F_0$ and $F_i$. By \eqref{eq:product}, $d_\infty(F,(G_0,\dots,G_M))\le\varepsilon$.
\end{proof}

\begin{remark}[Grid partitions suffice]\label{rem:grid}
The proofs of \cref{prop:real,prop:padic} only use the grid partitions $\{\cB_V(\bc,\bp^{-\bk}):\bc\in C_\bk\}$ of \cref{prop:partition}, in which all balls share one multi-radius and the centres range over $C_\bk$. \Cref{thm:main} therefore also holds for the smaller classes in which $\cP$ and $\cP_i$ are grid partitions.
\end{remark}

\section{Architecture}
\label{sec:architecture}

\subsection{A first adelic architecture}

\Cref{thm:main} holds for any dense families $\cF_\R$ and $\cF_{p_i}$. Its interest for machine learning comes from choosing them to be \emph{computable}. Every element of the resulting classes is then described by finitely many parameters and evaluated by finitely many elementary operations: affine maps, coordinatewise activations, and ball indicators. The indicators have a closed form (\cref{lem:indicator}), $\ind_{\cB_V(\bc,\bp^{-\bs})}(\bz)=\ind_{\Kp}\big(\bp^{-\bs}\odot(\bz-\bc)\big)$, and for $\bz,\bc\in\Kp$ they reduce to a congruence test, $\bz_i\equiv\bc_i\pmod{p_i^{s_i}}$ for all $i$, which only reads the first $s_i$ digits of each coordinate. Taking the perceptrons of \citet{cybenko1989approximation} and the character networks of \citet{mihara} thus turns \cref{thm:main} into explicit, computable universal families, which are natural candidates for adelic network architectures. Below, a ball $U\subset\Kp$ of $V$ is written $U=\cB_V(\bc_U,\bp^{-\bs_U})$ with $\bc_U\in\Kp$ and $\bs_U\in\N^M$, and similarly for balls of $\widehat V_i$ contained in $K_{\hp_i}$, with the primes $\hp_i$.

\begin{mdframed}[linewidth=0.8pt, linecolor=blue!50!black, backgroundcolor=blue!4, roundcorner=4pt, innertopmargin=4pt, innerbottommargin=8pt, innerleftmargin=8pt, innerrightmargin=8pt, nobreak=true]
\begin{corollary}\label{cor:explicit}
Let $\sigma:\R\to\R$ be continuous and sigmoidal, and for each $i$ let $\chi_i$ be an injective $p_i$-adic character as in \cref{thm:mihara}, all acting coordinatewise. Let
\[
  N_\R=\left\{
  \begin{aligned}
    f\colon K&\longrightarrow\R^n\\
    (\bx,\bz)&\longmapsto\sum_{U\in\cP}\ind_{\Kp}\big(\bp^{-\bs_U}\odot(\bz-\bc_U)\big)\,W_U\,\sigma(A_U\bx+\bb_U)\\
    &\textup{$\cP$ a finite partition of $\Kp$ into balls}
  \end{aligned}
  \ ,\ 
  \begin{aligned}
    &h_{U}\in\N,\ A_U\in\R^{h_{U}\times d_0}\\
    &\bb_U\in\R^{h_U},\ W_U\in\R^{n\times h_U}
  \end{aligned}
  \right\},
\]
and, for $i\in\{1,\dots,M\}$,
\[
  N_{p_i}=\left\{
  \begin{aligned}
    f\colon K&\longrightarrow\Q_{p_i}^{n_i}\\
    (\bx,\bz)&\longmapsto\sum_{U\in\cP_i}\ind_{K_{\hp_i}}\big(\hp_i^{-\bs_U}\odot(\hz_i-\bc_U)\big)\,C_U\,\chi_i(A_U\bz_i+\bb_U)\\
    &\textup{$\cP_i$ a finite partition of $K_{\hp_i}$ into balls}
  \end{aligned}
  ,\,
  \begin{aligned}
    &h_U\in\N,\ A_U\in\Zpi^{h_U\times d_i}\\
    &\bb_U\in\Zpi^{h_U},\ C_U\in\Qpi^{n_i\times h_U}
  \end{aligned}
  \right\}.
\]
Then $N_\R$ is dense in $\cC(K,\R^n)$ and, for every $i$, $N_{p_i}$ is dense in $\cC(K,\Q_{p_i}^{n_i})$ for the uniform distance. Consequently, $N_\R\times\prod_{i=1}^M N_{p_i}$ is dense in $\cC(K,\cY)$, and in particular in $\cC(K,\cX)$ when $n=d_0$ and $n_i=d_i$.
\end{corollary}
\end{mdframed}

\begin{proof}
$N_{\R}$ is $S_{\R}$ instantiated with $\cF_\R$ being the one-hidden-layer perceptrons which form a uniformly dense family of $\cC(\KR,\R^n)$ \citep{cybenko1989approximation}. $N_{p_{i}}$ is $S_{p_{i}}$ instantiated with $\cF_{p_i}$ being the character networks of \cref{thm:mihara} which also form a uniformly dense family of $\cC(\Zpi^{d_i},\Q_{p_i}^{n_i})$. With \cref{lem:indicator}, each indicator can be rewritten: $\ind_{\Kp}\big(\bp^{-\bs_U}\odot(\bz-\bc_U)\big) = \ind_{U}(\bz)$ and $\ind_{K_{\hp_i}}\big(\hp_i^{-\bs_U}\odot(\hz_i-\bc_U)\big) = \ind_{U}(\hz_i)$.\\
We conclude by applying \cref{thm:main}.
\end{proof}

\subsection{Analysis}

To discuss this architecture, we specialise to $\R\times\Q_p$ \textit{i.e.} $M=1$ and $d_0=d_1=n=n_1=1$, $K=\KR\times\Z_p$ with $\KR\subset\R$ a compact interval. The layer of \cref{cor:explicit} becomes, for a finite partition $\cP$ of $\Z_p$ into balls $U=\cB(c_U,p^{-s_U})$,
\[
  \hat F(x,z)=\sum_{U\in\cP}\ind_{\Z_p}\Big(\frac{z-c_U}{p^{s_U}}\Big)\mathbf{w}_U^{\top}\sigma(\mathbf{a}_Ux+\bb_U),
  \qquad
  \hat G(x,z)=\boldsymbol{\lambda}^{\top}\chi(\mathbf{a}z+\bb),
\]
with $\mathbf{w}_U,\mathbf{a}_U,\bb_U\in\R^{h_{U}}$, $\mathbf{a},\bb\in\Z_p^{h}$ and $\boldsymbol{\lambda}\in\Q_p^{h}$, the activations acting coordinatewise. Since $M=1$, the $p$-adic branch contains no indicator: it is a plain character network on $z$, and it ignores $x$. \Cref{fig:arch-RxQp} shows the general layer with two primes, where the ball indicators also couple the $p$-adic coordinates.

\begin{itemize}
  \item \emph{A $p$-adic mixture of experts with hard routing.} The real branch can be read as a mixture of real experts, one per ball of $\cP$, where the $p$-adic input acts as a deterministic router: $\ind_{\Z_p}((z-c_U)/p^{s_U})=1$ exactly when $z\equiv c_U\pmod{p^{s_U}}$, so exactly one expert is active for each $z$.
  \item \emph{Maybe a bit degenerate.} The number of balls grows with the resolution (a partition into balls of radius $p^{-k}$ has $p^k$ of them), the real networks share no parameters, and the dependence on $z$ is that of a lookup table over residues. The indicators are locally constant, so the output has zero derivative with respect to $z$, and the ball parameters ($c_U$, $s_U$) are discrete.
\end{itemize}

\subsection{Covers instead of partitions}
\label{sec:covers}

A first relaxation is to allow the balls to overlap. Let $S_\R^{\cup}$ be defined as $S_\R$ in \cref{thm:main}, but with $\cU$ a finite \emph{cover} of $\Kp$ by balls of $V$ instead of a partition:
\[
  (\bx,\bz)\longmapsto\sum_{U\in\cU}\ind_U(\bz)\,f_U(\bx),\qquad \Kp=\bigcup_{U\in\cU}U,\quad f_U\in\cF_\R.
\]
Every partition is a cover, so $S_\R\subset S_\R^{\cup}$ and $S_\R^{\cup}$ is dense in $\cC(K,\R^n)$ by \cref{thm:main}.

Several terms can now be nonzero at the same point. 
In the case $\R\times\Q_p$, two balls of $\Z_p$ are either disjoint or nested, so the balls containing $z$ form a chain $U_1\supsetneq U_2\supsetneq\dots$, and the output $\sum_{U\ni z}f_U(x)$ reads as a successive refinement along the branch of the $p$-adic tree that contains $z$. Larger balls give terms shared by many residue classes, and smaller balls add corrections. (For $M\ge2$, balls of $V$ that intersect need not be nested, since the radii can compare differently at different primes.) The same relaxation applies to the $p_i$-adic branches $S_{p_i}$ when $M\ge 2$: replacing the partition $\cP_i$ of $K_{\hp_i}$ by a finite ball cover $\cU_i$ defines a class $S_{p_i}^{\cup}\supset S_{p_i}$ that remains dense in $\cC(K,\Q_{p_i}^{n_i})$, since $\ind_U(\hz_i)\in\{0,1\}\subset\Q_{p_i}$. Still, this class feels somewhat degenerate: the dependence on the gating coordinates remains locally constant.

\subsection{Fourier expansion of the indicators}
\label{sec:fourier}

A second relaxation replaces the indicators by their Fourier expansion over the finite groups $\Zpi/p_i^{s_i}\Zpi$ (\cref{app:fourier}). For $\bs\in\N^M$, $\bgamma\in C_\bs$ and $\bu\in\Kp$, let
\[
  \theta_{\bgamma,\bs}(\bu)=\sum_{i=1}^M\Big\{\frac{\langle\bgamma_i,\bu_i\rangle}{p_i^{s_i}}\Big\}_{p_i}\in[0,M).
\]
By \cref{lem:fourier}, every ball $U=\cB_V(\bc_U,\bp^{-\bs_U})\subset\Kp$ satisfies, for $\bz\in\Kp$,
\[
  \ind_U(\bz)=\frac{1}{|C_{\bs_U}|}\sum_{\bgamma\in C_{\bs_U}}\cos\big(2\pi\,\theta_{\bgamma,\bs_U}(\bz-\bc_U)\big).
\]

\begin{proposition}\label{prop:cos}
Assume that $\cF_\R$ is stable under multiplication by positive scalars, as perceptrons are. Then the class $S_\R^{\cos}$ of maps
\[
  (\bx,\bz)\longmapsto\sum_{j=1}^J\cos\big(2\pi\,\theta_{\bgamma_j,\bs_j}(\bz-\bc_j)\big)\,f_j(\bx),
  \qquad J\in\N,\ \bs_j\in\N^M,\ \bgamma_j\in C_{\bs_j},\ \bc_j\in\Kp,\ f_j\in\cF_\R,
\]
is dense in $\cC(K,\R^n)$.
\end{proposition}

\begin{proof}
By \cref{thm:main}, $S_\R$ is dense. Write each ball of an element $\sum_{U\in\cP}\ind_U(\bz)f_U(\bx)$ of $S_\R$ as $U=\cB_V(\bc_U,\bp^{-\bs_U})$ and expand its indicator as above. This gives $\sum_{U\in\cP}\sum_{\bgamma\in C_{\bs_U}}\cos\big(2\pi\theta_{\bgamma,\bs_U}(\bz-\bc_U)\big)\,|C_{\bs_U}|^{-1}f_U(\bx)$, which belongs to $S_\R^{\cos}$ since $|C_{\bs_U}|^{-1}f_U\in\cF_\R$. Hence $S_\R\subset S_\R^{\cos}$.
\end{proof}

In the case $\R\times\Q_p$, an element of $S_\R^{\cos}$ reads $\sum_{j=1}^J\cos\big(2\pi\{\gamma_j(z-c_j)/p^{s_j}\}_p\big)\,\mathbf{w}_j^{\top}\sigma(\mathbf{a}_jx+\bb_j)$. Each real network is weighted by the cosine of a $p$-adic phase, so all terms contribute at once with weights in $[-1,1]$ that depend on $z$. Unlike the cover relaxation $S_\R^{\cup}$, this Fourier relaxation only applies to the real branch $S_\R$, because the real weights $\cos(2\pi\theta_{\bgamma,\bs}(\bz-\bc))\in[-1,1]\subset\R$ cannot multiply $\Q_{p_i}^{n_i}$-valued networks $g_U(\bz_i)$. Moreover, while the proof of \cref{prop:cos} expands each ball indicator into $|C_{\bs_U}|=\prod_{i=1}^M p_i^{s_{U,i}d_i}$ cosines (which would yield $|C_\bs|^2$ terms for a full grid partition of resolution $\bs$), in practice $S_\R^{\cos}$ is meant to be used with a small budget $J\ll\sum_{U\in\cP}|C_{\bs_U}|$ of sampled or learnt frequencies $(\bgamma_j,\bs_j)$ rather than full character sums.

\section{Discussion and Perspectives}
\label{sec:opening}

\paragraph{Theoretical scope.}
This paper is purely theoretical. As with any universal approximation theorem, density only shows that the proposed classes can represent arbitrary continuous maps given enough balls or frequencies; it says nothing about sample complexity, generalisation, or practical performance. Evaluating whether these architectures are useful in practice requires training them, which immediately raises the question of how to optimise their $p$-adic parameters.

\paragraph{Limits of current $\mathbf{p}$-adic optimisation.}
While \cref{cor:explicit,prop:cos} provide explicit, computable architectures, training their $p$-adic parameters does not fall directly within the continuous optimisation framework of \citet{salazar2026continuous}, which relies on lifting $p$-adic analytic functions to the Berkovich affine line via $\zeta \mapsto \abs{h}_\zeta$. 
The base $p_i$-adic networks are not the issue: for $\chi_i(x)=a_i^x$ with $a_i\in 1+p_i\Zpi$ ($1+4\Z_2$ for $p_i=2$), the map $\chi_i(A_U\bz_i+\bb_U)=\exp_{p_i}\big((\log_{p_i} a_i)(A_U\bz_i+\bb_U)\big)$ is a strictly convergent power series on $\Zpi$, so each expert $C_U\chi_i(A_U\bz_i+\bb_U)$ is globally analytic in its parameters $(A_U,\bb_U,C_U)$ (and for $M=1$, $N_{p_1}$ is already analytic). The difficulty comes from the gates:
\begin{itemize}
  \item \emph{Step-function gating in the real branch ($N_\R$).} Because $N_\R$ is real-valued, it cannot be a $p$-adic power series of its parameters; the natural class compatible with \citet{salazar2026continuous} consists of smooth real functions of Berkovich-liftable norms $\abs{h(\bz;\theta)}_{p_i}$. A ball indicator writes $\ind_{\cB(c,p^{-s})}(z)=H\big(p^{-s}-\abs{z-c}_p\big)$, where $z-c$ is linear in $c$ but $H=\ind_{[0,\infty)}$ is a real step function. The non-smoothness comes from $H$, which makes the gate locally constant with respect to the centres $\bc_U$ (while the radii $\bs_U$ are discrete on $\Q_{p_i}$, though they become continuous in Berkovich space below).
  \item \emph{Additive phases in the Fourier variant ($S_\R^{\cos}$).} Replacing indicators by cosines does not remove this obstacle: the phase $\theta_{\bgamma,\bs}(\bz-\bc)$ is locally constant in $\bc$ everywhere, the frequencies $\bgamma$ are discrete, and additive characters have no Berkovich lift.
  \item \emph{Cross-prime gating in the $p_i$-adic branch ($N_{p_i}$, $M\ge 2$).} When $M\ge 2$, the $p_i$-adic output is gated by balls in the \emph{other} primes $p_k\neq p_i$. By \cref{lem:constant}, any continuous map from a connected parameter space (such as the Berkovich line over $\Q_{p_k}$) to $\Q_{p_i}^{n_i}$ is constant, so cross-prime gate parameters cannot be relaxed continuously while keeping the output in $\Q_{p_i}^{n_i}$.
\end{itemize}

\paragraph{Connection with Berkovich space via $\mathbf{p}$-adic balls.}
Still, the parameters of our architectures have a direct interpretation in the Berkovich space of \citet{salazar2026continuous}. 
Across all three variants (\cref{sec:covers,cor:explicit,prop:cos}), each gate or phase depends on $(\bc_U, \bs_U)$ only through the closed ball $U = \cB_V(\bc_U, \bp^{-\bs_U})$ it defines: because every point of a $p$-adic ball is a centre (and shifting $\bc_i$ by an element of $p_i^{s_i}\Zpi^{d_i}$ adds a $p_i$-adic integer to $\langle\bgamma_i,\bz_i-\bc_i\rangle/p_i^{s_i}$, leaving its fractional part unchanged), replacing $\bc_U$ by any other $\bc'_U \in U$ leaves the model unchanged. 
In the one-dimensional case ($M=1$, $d_1=1$), closed balls $\cB(c, p^{-s}) \subset \Q_p$ are directly points of the Berkovich affine line into which \citet{salazar2026continuous} embed $\Q_p$ to perform continuous optimisation. 
Each ball $U$ can thus be viewed as a single point in Berkovich space, which appears here as the natural parameter space of the gates in its own right rather than a detour to perform optimisation. Extending this correspondence to multidimensional or multi-prime anisotropic balls $\cB_V(\bc_U,\bp^{-\bs_U})$ is more involved and is left for future work.

\paragraph{Toward training an adelic model.}
A natural next step would be to train a small model of this form on a toy task on $\R \times \Q_p$ that mixes continuous variations of magnitude with hierarchical or modular structure—for instance by training the real weights $(\mathbf{w}_U, \mathbf{a}_U, \bb_U)$ by gradient descent while updating the gating balls in Berkovich space or relaxing the cosine weights of \cref{sec:fourier}.

\bibliography{references}

\clearpage
\appendix

\section{Topological preliminaries}
\label[appsec]{app:compact}

\begin{lemma}\label{lem:compact}
The max metric $d$ induces the product topology on $\cX$, and $K=\KR\times\Kp$ is compact.
\end{lemma}

\begin{proof}
$d$ is the maximum of finitely many metrics on the factors, hence a metric. Its open balls are products of open balls of the factors, so it induces the product topology. $\Z_p$ is compact: it is complete and, by \cref{prop:partition}, covered for every $k$ by $p^k$ balls of radius $p^{-k}$, hence totally bounded. Thus each $\Zpi^{d_i}$ is compact, and $K$ is a finite product of compact spaces, hence compact.
\end{proof}

Heine's theorem then applies to every $F\in\cC(K,\cY)$ and every $\tilde F\in\cC(\Kp,\Q_{p_i}^{n_i})$.

\begin{proof}[Proof of \cref{lem:constant}]
Fix $x_0\in X$. For $k\in\N$, the ball $\cB_k=\cB(g(x_0),p^{-k})$ is clopen, so $g^{-1}(\cB_k)$ is clopen in $X$. It contains $x_0$, so by connectedness $g^{-1}(\cB_k)=X$. Hence $g(X)\subset\bigcap_k\cB_k=\{g(x_0)\}$.
\end{proof}

\section{The \texorpdfstring{$\R\times\Q_p$}{R x Qp} case}
\label[appsec]{app:demo}

This appendix treats the smallest cases with simple notation. It contains all the ideas of \cref{thm:main}, whose proof only adds bookkeeping for several primes and dimensions.

Which architecture should we look for? Proposing one directly would risk being somewhat arbitrary, so we look at the simplest mixed space and see what structure appears. Let $K=\KR\times\Z_p$ with $\KR\subset\R$ a compact interval, equipped with $d((x,z),(x',z'))=\max(\abs{x-x'},\abs{z-z'}_p)$. Let $\cF_\R$ and $\cF_p$ be dense in $\cC(\KR,\R)$ and $\cC(\Z_p,\Q_p)$.

\begin{proposition}\label{prop:demo-RQp}
\leavevmode
\begin{enumerate}[label=(\roman*)]
  \item \emph{No real variable in the $p$-adic coordinate.} Every $F\in\cC(K,\Q_p)$ satisfies $F(x,z)=F(x_0,z)$ for all $x,x_0\in\KR$ and $z\in\Z_p$. Hence the maps $(x,z)\mapsto g(z)$, $g\in\cF_p$, are dense in $\cC(K,\Q_p)$.
  \item \emph{Real coordinate.} The maps $(x,z)\mapsto\sum_{c=0}^{p^k-1}\ind_{\cB(c,p^{-k})}(z)\,f_c(x)$, with $k\in\N$ and $f_c\in\cF_\R$, are dense in $\cC(K,\R)$.
\end{enumerate}
\end{proposition}

\begin{proof}
(i) For fixed $z$, $x\mapsto F(x,z)$ is a continuous map from the connected interval $\KR$ to the totally disconnected space $\Q_p$, hence constant by \cref{lem:constant}. So $F(x,z)=\tilde F(z)$ with $\tilde F=F(x_0,\cdot)\in\cC(\Z_p,\Q_p)$, and any $g\in\cF_p$ with $\sup\abs{\tilde F-g}_p\le\varepsilon$ gives $\sup_K\abs{F(x,z)-g(z)}_p\le\varepsilon$.

(ii) \emph{Heine's theorem and a cut of $\Z_p$.} Let $F\in\cC(K,\R)$ and $\varepsilon>0$. $K$ is compact, so $F$ is uniformly continuous: there is $\eta>0$ with $\abs{F(x,z)-F(x',z')}\le\varepsilon/2$ whenever $d((x,z),(x',z'))\le\eta$. Choose $k$ with $p^{-k}\le\eta$. By \cref{prop:partition}, $\Z_p=\bigsqcup_{c=0}^{p^k-1}\cB(c,p^{-k})$. For $z\in\cB(c,p^{-k})$, $d((x,z),(x,c))=\abs{z-c}_p\le\eta$, so $\abs{F(x,z)-F(x,c)}\le\varepsilon/2$ for all $x$. For all $c$, $F(\cdot,c) \in \cC(\KR,\R)$. Therefore, there exists $f_c\in\cF_\R$ with $\sup_{\KR}\abs{F(\cdot,c)-f_c}\le\varepsilon/2$. Then $G(x,z)=\sum_c\ind_{\cB(c,p^{-k})}(z)f_c(x)$ satisfies $\abs{F(x,z)-G(x,z)}\le\varepsilon$ on $K$, since exactly one ball contains $z$.
\end{proof}

The resulting model is a sum of real networks in $x$, each multiplied by the indicator of a residue class of $z$ modulo $p^k$.

\paragraph{The case $\Q_{p_1}\times\Q_{p_2}$.}
Let $K=\Z_{p_1}\times\Z_{p_2}$ with $p_1\neq p_2$, and no real factor.

\begin{proposition}\label{prop:demo-QpQp}
The maps $(z_1,z_2)\mapsto\sum_{c=0}^{p_2^k-1}\ind_{\cB(c,p_2^{-k})}(z_2)\,g_c(z_1)$, with $k\in\N$ and $g_c\in\cF_{p_1}$, are dense in $\cC(K,\Q_{p_1})$.
\end{proposition}

\begin{proof}
Let $F\in\cC(K,\Q_{p_1})$ and $\varepsilon>0$. By uniform continuity, there is $k$ such that $\abs{F(z_1,z_2)-F(z_1,c)}_{p_1}\le\varepsilon$ whenever $z_2\in\cB(c,p_2^{-k})$. Pick $g_c\in\cF_{p_1}$ with $\sup\abs{F(\cdot,c)-g_c}_{p_1}\le\varepsilon$. By the ultrametric inequality, $\abs{F(z_1,z_2)-g_c(z_1)}_{p_1} = \abs{F(z_1,z_2)-F(z_{1}, c) + F(z_{1},c) - g_c(z_1)}_{p_1}\le\max(\abs{F(z_1,z_2)-F(z_{1}, c)}_{p_{1}}, \abs{F(z_{1},c) - g_c(z_1)}_{p_1})\le\max(\varepsilon,\varepsilon)=\varepsilon$ for $z_2\in\cB(c,p_2^{-k})$: the two errors combine through a maximum rather than a sum.
\end{proof}

\paragraph{Take-away.}
For a given output, each input coordinate plays one of three roles:
\begin{center}
\begin{tabular}{lccc}
  \toprule
  output $\backslash$ input & $\bx\in\R^{d_0}$ & $\bz_i\in\Q_{p_i}^{d_i}$ & $\bz_k\in\Q_{p_k}^{d_k}$, $k\ne i$\\
  \midrule
  $\R^n$ & processed by $\cF_\R$ & partitioned & partitioned\\
  $\Q_{p_i}^{n_i}$ & removed (connectedness) & processed by $\cF_{p_i}$ & partitioned\\
  \bottomrule
\end{tabular}
\end{center}
Here \emph{removed} means the output does not depend on the coordinate, \emph{partitioned} means the coordinate is frozen on the balls of a finite partition, and \emph{processed} means the coordinate is fed to a base approximator. The general proofs of \cref{sec:results} apply these mechanisms blockwise, and once the two cases above are understood, the rest is mostly bookkeeping.

\section{On multidimensional \texorpdfstring{$p$}{p}-adic balls}
\label[appsec]{app:balls}

\paragraph{Partition property.}

\begin{proof}[Proof of \cref{prop:partition}]
(i) For $\bc\in\Z_p^d$, $\norm{\bz-\bc}_p\le p^{-k}$ if and only if $\bz-\bc\in p^k\Z_p^d$, so $\cB(\bc,p^{-k})=\bc+p^k\Z_p^d$ is a coset of the subgroup $p^k\Z_p^d$. Since $\Z_p/p^k\Z_p\cong\Z/p^k\Z$, the set $C_k$ is a system of representatives of $\Z_p^d/p^k\Z_p^d$, and the cosets partition $\Z_p^d$.
(ii) Taking the product over $i$ of the partitions in (i),
\[
  \Kp=\prod_{i=1}^M\bigsqcup_{\bc_i\in C_{k_i}}\cB(\bc_i,p_i^{-k_i})
  =\bigsqcup_{\bc\in C_\bk}\prod_{i=1}^M\cB(\bc_i,p_i^{-k_i})
  =\bigsqcup_{\bc\in C_\bk}\cB_V(\bc,\bp^{-\bk}).
\]
(iii) If $\bz,\bz'\in\cB_V(\bc,\bp^{-\bk})$, then $\norm{\bz_i-\bz'_i}_{p_i}\le p_i^{-k_i}$ for every $i$, so $d_V(\bz,\bz')\le\max_ip_i^{-k_i}$.
\end{proof}

\paragraph{Indicator functions.}

\begin{lemma}[Indicators]\label{lem:indicator}
For $\bc\in\Q_p^d$ and $s\in\Z$, $\ind_{\cB(\bc,p^{-s})}(\bz)=\ind_{\Z_p^d}\big(p^{-s}(\bz-\bc)\big)=\ind_{p^s\Z_p^d}\big(\bz-\bc\big)$. Consequently, for $\bc\in V$ and $\bs\in\Z^M$,
\[
  \ind_{\cB_V(\bc,\bp^{-\bs})}(\bz)
  =\prod_{i=1}^M\ind_{\Zpi^{d_i}}\big(p_i^{-s_i}(\bz_i-\bc_i)\big)
  =\ind_{\Kp}\big(\bp^{-\bs}\odot(\bz-\bc)\big).
\]
\end{lemma}

\begin{proof}
$\norm{\bz-\bc}_p\le p^{-s}$ if and only if $\norm{p^{-s}(\bz-\bc)}_p=p^{s}\norm{\bz-\bc}_p\le1$, i.e.\ $p^{-s}(\bz-\bc)\in\Z_p^d$. The product formula follows because the indicator of a product of sets is the product of the indicators.
\end{proof}

For $\bc\in C_\bk$, the indicator of $\cB_V(\bc,\bp^{-\bk})$ equals $1$ exactly when $\bz_i\equiv\bc_i\pmod{p_i^{k_i}}$ for all $i$: it is a lookup of residues modulo prime powers.

\section{Fourier transform of indicators over \texorpdfstring{$\Q_p$}{Qp}}
\label[appsec]{app:fourier}

Throughout, $\frp{x}$ is the fractional part of $x\in\Q_p$ (\cref{sec:background}). We use two of its properties: for an integer $a$, $\{a/p^k\}_p=(a\bmod p^k)/p^k$; and $\frp{x+y}-\frp{x}-\frp{y}\in\Z$, since this number lies in $\Z_p\cap\Z[1/p]=\Z$.

\paragraph{The case $\Z_p$.}

\begin{lemma}\label{lem:fourier-Zp}
Let $k\in\N$. For $u\in\Z_p$,
\[
  \ind_{p^k\Z_p}(u)=\frac{1}{p^{k}}\sum_{\gamma=0}^{p^k-1}e^{2 \mathrm{i} \pi \,\{\gamma u/p^k\}_p}=\frac{1}{p^{k}}\sum_{\gamma=0}^{p^k-1}\cos\Big(2\pi\Big\{\frac{\gamma u}{p^k}\Big\}_p\Big).
\]
Consequently, for $c,z\in\Z_p$, $\ind_{\cB(c,p^{-k})}(z)=p^{-k}\sum_{\gamma=0}^{p^k-1}\cos\big(2\pi\{\gamma(z-c)/p^k\}_p\big)$.
\end{lemma}

\begin{proof}
Let $a\in\{0,\dots,p^k-1\}$ with $u\equiv a\pmod{p^k}$. Then $u = a + vp^k$ with $v \in \Z_p$ and for any $\gamma \in \{0,\dots,p^k-1\}$ $\frac{\gamma u}{p^k} = \frac{\gamma a}{p^k} + \gamma v$. 
Since $\gamma v \in \Z_{p}$ we have $\{\frac{\gamma u}{p^k}\}_p=\{\frac{\gamma a}{p^k}\}_p$. Also, $\gamma a \in \Z$ so the euclidean division of $\gamma a$ mod $p^k$ gives $\gamma a = r + qp^k$ with $r \in \{0,\dots,p^k-1\}$ and $q \in \Z$, which gives $\frac{\gamma a}{p^k} = \frac{r}{p^k} + q$ with $q \in \Z \subset \Z_{p}$ and $\frac{r}{p^k} \in \Q \cap [0,1)$ thus $\{\frac{\gamma u}{p^k}\}_p = \{\frac{\gamma a}{p^k}\}_p = \frac{r}{p^k}= \frac{\gamma a}{p^k} - q$ and so $\{\frac{\gamma u}{p^k}\}_p = \frac{\gamma a}{p^k} - q$ with $q \in \Z$. 
With this we can write $e^{2 \mathrm{i} \pi \,\{\gamma u/p^k\}_p} = e^{2 \mathrm{i} \pi \,\gamma a/p^k} = \omega^\gamma$  with $\omega = e^{2 \mathrm{i} \pi \,a/p^k}$. We can rewrite the sum $\sum_{\gamma=0}^{p^k-1}e^{2 \mathrm{i} \pi \,\{\gamma u/p^k\}_p} = \sum_{\gamma=0}^{p^k-1}\omega^\gamma$. If $u \in p^k\Z_{p}$ then $a=0$ and $\omega = 1$ thus $\sum_{\gamma=0}^{p^k-1}e^{2 \mathrm{i} \pi \,\{\gamma u/p^k\}_p} = p^k$. If $u \notin p^k\Z_{p}$ then $a>0$ and $\omega^{p^k} = 1$ so $\sum_{\gamma=0}^{p^k-1}e^{2 \mathrm{i} \pi \,\{\gamma u/p^k\}_p} = (\omega^{p^k} - 1)/(\omega-1)$ = 0.

\noindent We thus have $ \ind_{p^k\Z_p}(u)=\frac{1}{p^{k}}\sum_{\gamma=0}^{p^k-1}e^{2 \mathrm{i} \pi \,\{\gamma u/p^k\}_p}$ and since $\ind_{p^k\Z_p}(u) \in \R$ the imaginary parts of the exponentials must sum to zero. Thus the equality: $\ind_{p^k\Z_p}(u) = p^{-k}\sum_{\gamma=0}^{p^k-1}\cos\big(2\pi\{\gamma u /p^k\}_p\big)$. By \cref{lem:indicator}: $\ind_{\cB(c,p^{-k})}(z) = \ind_{p^k\Z_p}(z-c)=p^{-k}\sum_{\gamma=0}^{p^k-1}\cos\big(2\pi\{\gamma(z-c)/p^k\}_p\big)$.
\end{proof}

\paragraph{The case $\Z_p^d$.}

\begin{lemma}\label{lem:fourier-Zpd}
Let $k\in\N$ and $C_k=\{0,\dots,p^k-1\}^d$. For $\bu\in\Z_p^d$,
\[
  \ind_{p^k\Z_p^d}(\bu)=\frac{1}{p^{kd}}\sum_{\bgamma\in C_k}e^{2 \mathrm{i} \pi \,\{\langle\bgamma,\bu\rangle/p^k\}_p}=\frac{1}{p^{kd}}\sum_{\bgamma\in C_k}\cos\Big(2\pi\Big\{\frac{\langle\bgamma,\bu\rangle}{p^k}\Big\}_p\Big).
\]
Consequently, for $\bc,\bz\in\Z_p^d$, $\ind_{\cB(\bc,p^{-k})}(\bz)=p^{-kd}\sum_{\bgamma\in C_k}\cos\big(2\pi\{\langle\bgamma,\bz-\bc\rangle/p^k\}_p\big)$.
\end{lemma}

\begin{proof}
For the max norm, $\bu\in p^k\Z_p^d$ if and only if $u_l\in p^k\Z_p$ for every $l$, so $\ind_{p^k\Z_p^d}(\bu)=\prod_{l=1}^d\ind_{p^k\Z_p}(u_l)$. Expanding each factor with \cref{lem:fourier-Zp} gives
\[
\begin{aligned}
  \ind_{p^k\Z_p^d}(\bu)
  & = \frac{1}{p^{kd}} \prod_{l=1}^d\sum_{\gamma_{l} =0}^{p^k-1} \exp \left( 2 \mathrm{i}  \pi \left\{ \frac{\gamma_{l}u_{l}}{p^k} \right\}_p  \right) =
  \frac{1}{p^{kd}}\sum_{\bgamma \in C_k}\prod_{l=1}^d \exp \left( 2 \mathrm{i}  \pi \left\{ \frac{\gamma_{l}u_{l}}{p^k} \right\}_p  \right) \\
  & = \frac{1}{p^{kd}}\sum_{\bgamma \in C_k} \exp \left( 2 \mathrm{i}  \pi \sum_{l=1}^d\left\{ \frac{\gamma_{l}u_{l}}{p^k} \right\}_p  \right)
\end{aligned}
\]
On the other hand $\frp{x+y}-\frp{x}-\frp{y}\in\Z$ generalises to $\frp{\sum_{l=1}^dx_{l}} - \sum_{l=1}^d\frp{x_{l}} \in \Z$. Thus $\frp{\langle\bgamma,\bu\rangle/p^k} = \sum_{l=1}^d\frp{\frac{\gamma_{l}u_{l}}{p^k}} + m$ with $m \in \Z$. So $\exp(2 \mathrm{i} \pi \frp{\langle\bgamma,\bu\rangle/p^k}) = \exp(2 \mathrm{i} \pi \sum_{l=1}^d\frp{\frac{\gamma_{l}u_{l}}{p^k}})$ which completes the equality. Taking real parts gives the cosine form, and the ball case follows as in \cref{lem:fourier-Zp}.
\end{proof}

\paragraph{The case $\Kp$.}

\begin{lemma}\label{lem:fourier}
For $\bs\in\N^M$ and $\bc,\bz\in\Kp$, with $C_\bs$ as in \cref{prop:partition} and $\theta_{\bgamma,\bs}$ as in \cref{sec:fourier},
\[
  \ind_{\cB_V(\bc,\bp^{-\bs})}(\bz)=\frac{1}{|C_\bs|}\sum_{\bgamma\in C_\bs}e^{2 \mathrm{i} \pi \,\theta_{\bgamma,\bs}(\bz-\bc)}=\frac{1}{|C_\bs|}\sum_{\bgamma\in C_\bs}\cos\big(2\pi\,\theta_{\bgamma,\bs}(\bz-\bc)\big).
\]
\end{lemma}

\begin{proof}
By definition of the balls of $V$, $\cB_V(\bc,\bp^{-\bs})=\prod_{i=1}^M\cB(\bc_i,p_i^{-s_i})$, \\
so $\ind_{\cB_V(\bc,\bp^{-\bs})}(\bz)=\prod_{i=1}^M\ind_{\cB(\bc_i,p_i^{-s_i})}(\bz_i)$ with $\bc_i,\bz_i\in\Zpi^{d_i}$. Let $C_{i,s_i}=\{0,\dots,p_i^{s_i}-1\}^{d_i}$, so that $C_\bs=\prod_{i=1}^M C_{i,s_i}$ and $|C_\bs|=\prod_{i=1}^M p_i^{s_i d_i}$. Expanding each factor with \cref{lem:fourier-Zpd} (at the prime $p_i$ with $k=s_i$ and $\bu=\bz_i-\bc_i\in\Zpi^{d_i}$) gives
\[
\begin{aligned}
  \ind_{\cB_V(\bc,\bp^{-\bs})}(\bz)
  &= \prod_{i=1}^M \left( \frac{1}{p_i^{s_i d_i}} \sum_{\bgamma_i \in C_{i,s_i}} \exp \left( 2 \mathrm{i} \pi \left\{ \frac{\langle \bgamma_i, \bz_i - \bc_i \rangle}{p_i^{s_i}} \right\}_{p_i} \right) \right) \\
  &= \left( \prod_{i=1}^M \frac{1}{p_i^{s_i d_i}} \right) \sum_{\bgamma \in C_\bs} \prod_{i=1}^M \exp \left( 2 \mathrm{i} \pi \left\{ \frac{\langle \bgamma_i, \bz_i - \bc_i \rangle}{p_i^{s_i}} \right\}_{p_i} \right) \\
  &= \frac{1}{|C_\bs|} \sum_{\bgamma \in C_\bs} \exp \left( 2 \mathrm{i} \pi \sum_{i=1}^M \left\{ \frac{\langle \bgamma_i, \bz_i - \bc_i \rangle}{p_i^{s_i}} \right\}_{p_i} \right).
\end{aligned}
\]
By definition (\cref{sec:fourier}), $\theta_{\bgamma,\bs}(\bz-\bc) = \sum_{i=1}^M \big\{ \frac{\langle\bgamma_i,\bz_i-\bc_i\rangle}{p_i^{s_i}} \big\}_{p_i}$, which yields
\[
  \ind_{\cB_V(\bc,\bp^{-\bs})}(\bz) = \frac{1}{|C_\bs|}\sum_{\bgamma\in C_\bs}e^{2\mathrm{i}\pi\,\theta_{\bgamma,\bs}(\bz-\bc)}.
\]
Since $\ind_{\cB_V(\bc,\bp^{-\bs})}(\bz)\in\R$, the imaginary parts of the exponentials sum to zero, and taking real parts gives the cosine form.
\end{proof}

\section{Different input and output primes}
\label[appsec]{app:different_primes}

In \cref{sec:notations,sec:results}, the output space $\cY=\R^n\times\prod_{i=1}^M\Q_{p_i}^{n_i}$ is indexed by the same primes $(p_1,\dots,p_M)$ as the input space $\cX=\R^{d_0}\times\prod_{i=1}^M\Q_{p_i}^{d_i}$. Having different primes in the input and output reduces directly to this setting by taking $(p_1,\dots,p_M)$ to be the union of all input and output primes and allowing $d_i=0$ or $n_i=0$:
\begin{enumerate}[label=(\roman*)]
  \item \emph{Output prime absent from the input ($q\notin\{p_1,\dots,p_M\}$).} An output factor $\Q_q^{n_q}$ whose prime $q$ does not appear in the input is equivalent to including $q$ among $\{p_1,\dots,p_M\}$, say $q=p_1$ with input dimension $d_1=0$. Then $\Z_{p_1}^{d_1}=\{0\}$ is reduced to a point and $K_{\hp_1}=\Kp$. In \cref{cor:explicit}, the linear term $A_U\bz_1+\bb_U$ of the character network of \citet{mihara} reduces to the bias $\bb_U$, so each expert $C_U\chi_1(A_U\bz_1+\bb_U)=C_U\chi_1(\bb_U)$ naturally becomes a constant vector $\mathbf{v}_U\in\Q_{p_1}^{n_1}$, and $N_{p_1}$ consists of piecewise-constant maps on ball partitions of $\Kp$:
  \[
    (\bx,\bz)\longmapsto\sum_{U\in\cP_1}\ind_U(\bz)\,\mathbf{v}_U,\qquad \mathbf{v}_U\in\Q_{p_1}^{n_1}.
  \]
  This density of piecewise-constant maps can also be proved directly as in \cref{cor:no-real,prop:padic}: any $F\in\cC(K,\Q_q^{n_q})$ is independent of $\bx\in\KR$ by connectedness (\cref{lem:constant}) and uniformly continuous on the compact space $\Kp$, so partitioning $\Kp$ into balls $U$ of small enough radius and setting $\mathbf{v}_U=F(\bx_0,\bc_U)$ gives a uniform approximation of $F$.
  \item \emph{Input prime absent from the output.} Symmetrically, if an input prime $p_k$ has no corresponding output factor, setting $n_k=0$ omits the $p_k$-adic output branch $S_{p_k}$ while keeping the input coordinate $\bz_k\in\Z_{p_k}^{d_k}$ in the gating partitions $\cP$ and $\cP_i$ ($i\ne k$).
\end{enumerate}

\clearpage

\section{Index of notation}
\label[appsec]{app:index}

\begingroup
\renewcommand{\arraystretch}{1.15}
\begin{longtable}{@{}p{0.23\linewidth}p{0.55\linewidth}l@{}}
  \toprule
  Symbol & Meaning & Introduced in\\
  \midrule
  \endfirsthead
  \toprule
  Symbol & Meaning & Introduced in\\
  \midrule
  \endhead
  \bottomrule
  \endfoot
  \multicolumn{3}{@{}l}{\emph{Numbers and $p$-adic analysis}}\\
  $p$, $p_1,\dots,p_M$ & a prime; $M$ distinct primes & \cref{sec:notations}\\
  $\abs{\cdot}_p$ & $p$-adic absolute value & \cref{sec:background}\\
  $\Q_p$, $\Z_p$ & $p$-adic numbers, $p$-adic integers & \cref{sec:background}\\
  $\frp{x}$ & fractional part of $x\in\Q_p$ & \cref{sec:background}\\
  $\norm{\cdot}_p$ & max norm on $\Q_p^d$ & \cref{sec:background}\\
  $\norm{\cdot}$ & a norm on $\R^{d_0}$ & \cref{sec:notations}\\
  $\A_\Q$ & adele ring of $\Q$ & \cref{sec:background}\\
  \midrule
  \multicolumn{3}{@{}l}{\emph{Dimensions and multi-indices}}\\
  $M$ & number of primes & \cref{sec:notations}\\
  $d_0$, $d_i$ & real input dimension; dimension at the prime $p_i$ & \cref{sec:notations}\\
  $n$, $n_i$ & real output dimension; output dimension at $p_i$ & \cref{sec:notations}\\
  $\bp$ & $(p_1,\dots,p_M)$ & \cref{sec:notations}\\
  $\hp_i$ & $\bp$ without $p_i$, i.e.\ $(p_k)_{k\ne i}$ & \cref{sec:notations}\\
  $\bs$, $\bk$ & multi-indices in $\Z^M$, resp.\ $\N^M$ & \cref{sec:notations}\\
  $\bp^{\bs}\odot\bz$ & $(p_1^{s_1}\bz_1,\dots,p_M^{s_M}\bz_M)$ & \cref{sec:notations}\\
  \midrule
  \multicolumn{3}{@{}l}{\emph{Spaces, points and metrics}}\\
  $V$ & $\prod_{i=1}^M\Q_{p_i}^{d_i}$, the $p$-adic part of the space & \cref{sec:notations}\\
  $\widehat V_i$ & $\prod_{k\ne i}\Q_{p_k}^{d_k}$ & \cref{sec:notations}\\
  $\cX$ & proxi-adelic space $\R^{d_0}\times V$ & \cref{sec:notations}\\
  $\cY$ & output space $\R^n\times\prod_{i}\Q_{p_i}^{n_i}$ & \cref{sec:notations}\\
  $(\bx,\bz)$ & point of $\cX$: real part $\bx\in\R^{d_0}$, $p$-adic part $\bz\in V$ & \cref{sec:notations}\\
  $\bz_i$ & $i$-th block of $\bz$, in $\Q_{p_i}^{d_i}$ & \cref{sec:notations}\\
  $\hz_i$ & $\bz$ without its $i$-th block, in $\widehat V_i$; $\bz=(\bz_i,\hz_i)$ & \cref{sec:notations}\\
  $d_V$, $d_{\widehat V_i}$ & max metrics on $V$, $\widehat V_i$ & \cref{sec:notations}\\
  $d$, $d_\cY$ & max metrics on $\cX$, $\cY$ & \cref{sec:notations}\\
  \midrule
  \multicolumn{3}{@{}l}{\emph{Balls and partitions}}\\
  $\cB(\bc,r)$ & closed ball of $\Q_p^d$ of centre $\bc$ and radius $r$ & \cref{sec:background}\\
  $\cB_V(\bc,\bp^{-\bs})$ & anisotropic ball $\prod_i\cB(\bc_i,p_i^{-s_i})$ of $V$ & \cref{def:balls}\\
  $C_k$ & $\{0,\dots,p^k-1\}^d$, centres of the balls of radius $p^{-k}$ in $\Z_p^d$ & \cref{prop:partition}\\
  $C_\bk$ & $\prod_i\{0,\dots,p_i^{k_i}-1\}^{d_i}$, centres of a grid partition of $\Kp$ & \cref{prop:partition}\\
  $\cP$, $\cP_i$ & finite partitions of $\Kp$, $K_{\hp_i}$ into balls & \cref{thm:main}\\
  $\cU$ & finite cover of $\Kp$ by balls & \cref{sec:covers}\\
  $\bc_U$, $\bs_U$ & a centre and the multi-radius exponent of a ball $U$ & \cref{cor:explicit}\\
  \midrule
  \multicolumn{3}{@{}l}{\emph{Domains}}\\
  $\KR$ & non-empty compact connected subset of $\R^{d_0}$ & \cref{sec:notations}\\
  $\Kp$ & $\prod_{i=1}^M\Zpi^{d_i}$ & \cref{sec:notations}\\
  $K_{\hp_i}$ & $\prod_{k\ne i}\Z_{p_k}^{d_k}$ (a point if $M=1$) & \cref{sec:notations}\\
  $K$ & $\KR\times\Kp$, the compact domain & \cref{sec:notations}\\
  \midrule
  \multicolumn{3}{@{}l}{\emph{Function spaces and approximating classes}}\\
  $\cC(A,B)$ & continuous maps $A\to B$ & \cref{sec:notations}\\
  $d_\infty$ & uniform distance & \cref{sec:notations}\\
  $\cF_\R$ & dense subset of $\cC(\KR,\R^n)$ & \cref{sec:notations}\\
  $\cF_{p_i}$ & dense subset of $\cC(\Zpi^{d_i},\Q_{p_i}^{n_i})$ & \cref{sec:notations}\\
  $S_\R$, $S_{p_i}$ & approximating classes of the main theorem & \cref{thm:main}\\
  $N_\R$, $N_{p_i}$ & $S_\R$, $S_{p_i}$ instantiated with real perceptrons, resp.\ $p_i$-adic character networks & \cref{cor:explicit}\\
  $S_\R^{\cup}$, $S_{p_i}^{\cup}$ & variants of $S_\R$, $S_{p_i}$ with ball covers & \cref{sec:covers}\\
  $S_\R^{\cos}$ & variant of $S_\R$ with cosine modulations & \cref{prop:cos}\\
  $\theta_{\bgamma,\bs}$ & phase $\theta_{\bgamma,\bs}(\bu)=\sum_i\{\langle\bgamma_i,\bu_i\rangle/p_i^{s_i}\}_{p_i}$, for $\bgamma\in C_\bs$ and $\bu=(\bu_1,\dots,\bu_M)\in\Kp$ & \cref{sec:fourier}\\
  \midrule
  \multicolumn{3}{@{}l}{\emph{Activations}}\\
  $\sigma$ & real activation (sigmoidal or non-polynomial) & \cref{sec:background}\\
  $\chi$, $\chi_i$ & injective $p$-adic character $\Z_p\to\Q_p^\times$, resp.\ at $p_i$ & \cref{sec:background}\\
\end{longtable}
\endgroup

\end{document}